\documentclass{elsarticle}
\usepackage{xcolor}
\usepackage{graphicx}
\usepackage{amsmath}   
\usepackage{amsfonts}  
\usepackage{amssymb}   
\usepackage{multirow} 
\usepackage{amssymb}
\usepackage{csquotes}
\usepackage{pifont}
\usepackage{hyperref}  

\begin{document}
\begin{frontmatter}
\title{Anatomy-Guided Representation Learning Using a Transformer-Based Network for Thyroid Nodule Segmentation in Ultrasound Images}
\author{Muhammad Umar Farooq$^{1}$, Abd Ur Rehman$^{2}$, Azka Rehman$^{3}$, Muhammad Usman$^{4}$,Dong-Kyu Chae$^{5}$, Junaid Qadir$^{6}$}

\address{%
$^{1}$ \quad Department of Computer Science, Hanyang University, Seoul, 04762, South Korea  \\}
\address{%
$^{2}$ \quad Department of Computer Science, The University of Alabama, Seoul, 04762, South Korea \\}
\address{%
$^{3}$ \quad Department of Biomedical Sciences, Seoul National University, Seoul, 08826, South Korea (azkarehman@snu.ac.kr) \\}
\address{%
$^{4}$\quad Department of Anesthesiology, Perioperative and Pain Medicine, Stanford University, CA 94305, USA (usmanm@stanford.edu)}
\address{%
$^{5}$ \quad Department of Computer Science, Hanyang University, Seoul, 04762, South Korea (dongkyu@hanyang.ac.kr)} 
\address{%
$^{6}$ \quad Department of Computer Engineering, Qatar University, Doha, Qatar (jqadir@qu.edu.qa) \\}


\begin{abstract}

Accurate thyroid nodule segmentation in ultrasound images is critical for diagnosis and treatment planning. However, ambiguous boundaries between nodules and surrounding tissues, size variations, and the scarcity of annotated ultrasound data pose significant challenges for automated segmentation. Existing deep learning models struggle to incorporate contextual information from the thyroid gland and generalize effectively across diverse cases. To address these challenges, we propose SSMT-Net, a Semi-Supervised Multi-Task Transformer-based Network that leverages unlabeled data to enhance Transformer-centric encoder feature extraction capability in an initial unsupervised phase. In the supervised phase, the model jointly optimizes nodule segmentation, gland segmentation, and nodule size estimation, integrating both local and global contextual features. Extensive evaluations on the TN3K and DDTI datasets demonstrate that SSMT-Net outperforms state-of-the-art methods, with higher accuracy and robustness, indicating its potential for real-world clinical applications.


\end{abstract}
\end{frontmatter}

\section{Introduction}
\label{sec:intro}

Automated thyroid nodule segmentation in ultrasound imaging plays a pivotal role in supporting radiologists by improving diagnostic accuracy and reducing inter-observer variability \cite{radhachandran2024multitask}. Despite its importance, the task remains challenging due to heterogeneous echogenic patterns, ambiguous boundaries, and strong acoustic shadows commonly present in thyroid ultrasound scans \cite{li2023novel}. While deep learning-based segmentation frameworks have demonstrated promising performance \cite{liu2020end}, their ability to capture long-range dependencies—critical for accurate nodule delineation—remains limited \cite{song2024survey,farooq2025gdssa}. These limitations become more pronounced in cases with substantial variations in nodule morphology.

Existing approaches (e.g., \cite{zhang2024shan,gong2023thyroid}) predominantly rely on single-task learning, often overlooking complementary contextual cues such as thyroid gland structure, gland–nodule spatial interactions, and morphological priors. At the same time, the field faces a persistent shortage of large-scale annotated datasets, restricting model generalizability in real-world conditions. In contrast, recent advances in medical imaging demonstrate the benefits of incorporating auxiliary tasks and attention-driven architectures across several domains, such as mandibular canal delineation \cite{usman2022dual}, lung nodule segmentation with adaptive ROI selection \cite{usman2023deha,usman2020volumetric,usman2025multi}, brain tumor segmentation \cite{ullah2022cascade,rehman2023selective}, diabetic retinopathy segmentation \cite{ullah2023ssmd}, and broader biomedical detection challenges \cite{iqbal2023ldmres,ullah2023densely,ullah2023mtss}. These works consistently highlight how multi-scale attention, ROI adaptation, and multi-encoder feature fusion contribute to improved lesion localization and robustness.

Transformer-based architectures have further accelerated progress by enabling superior long-range context modeling, benefiting not only medical image reconstruction \cite{latif2018automating,usman2020retrospective,usman2024advancing,rehman2024biological,usmancomplex} but also segmentation and detection tasks across MRI, CBCT, and CT modalities. Their ability to integrate global and local representations has proven effective in diverse clinical workflows, such as cardiomegaly assessment \cite{lee2021evaluation}, multimodal neuroimaging fusion \cite{usman2024meds,latif2020leveraging}, phonocardiographic signal analysis \cite{latif2018phonocardiographic}, and cross-lingual feature representation learning \cite{latif2018cross}. Collectively, these studies demonstrate that multi-task learning (MTL), multi-encoder architectures, and attention-rich designs consistently outperform strictly single-task CNN systems, especially in complex and low-contrast settings.

MTL in particular has shown strong potential in enhancing primary task performance by leveraging the inductive bias of complementary auxiliary tasks \cite{karkalousos2024atommic,vandenhende2021multi}. In medical imaging, MTL-driven designs have yielded notable improvements in lung nodule detection \cite{usman2024meds}, metaverse-based brain-age estimation \cite{usman2024advancing}, COVID-19 classification \cite{ullah2023densely,ullah2023mtss}, and other clinical prediction tasks \cite{usman2017using,latif2018mobile}. Semi-supervised strategies also continue to play an important role in addressing data scarcity by effectively utilizing unlabeled samples for model pretraining or auxiliary task optimization \cite{latif2018automating,usman2020retrospective,usman2023deha}.

Motivated by these advances, we propose SSMT-Net, a Semi-Supervised Multitask Transformer-based Network designed specifically for thyroid nodule segmentation. Our framework combines supervised auxiliary tasks—gland segmentation and nodule size prediction—with unsupervised image reconstruction to mitigate limited annotated data and enrich feature representation learning. The model first undergoes self-supervised pretraining on unlabeled ultrasound scans to learn robust structural and textural priors, followed by supervised joint optimization across the auxiliary and primary tasks. The transformer-based encoder-decoder facilitates enhanced global context modeling, while multi-task learning acts as a regularizer that improves the network’s ability to differentiate subtle boundary variations.

Comprehensive evaluations on the TN3K and DDTI datasets demonstrate that SSMT-Net achieves state-of-the-art performance in accuracy and robustness. By integrating multi-task supervision, semi-supervised learning, transformer-based global reasoning, and clinically meaningful auxiliary objectives, our method offers a reliable and generalizable solution for real-world thyroid nodule segmentation workflows.

\section{Method}
\label{sec:method}
\subsection{Overall Architecture}

\begin{figure*}[!ht]
\centering
\centerline{\includegraphics[width=\textwidth]{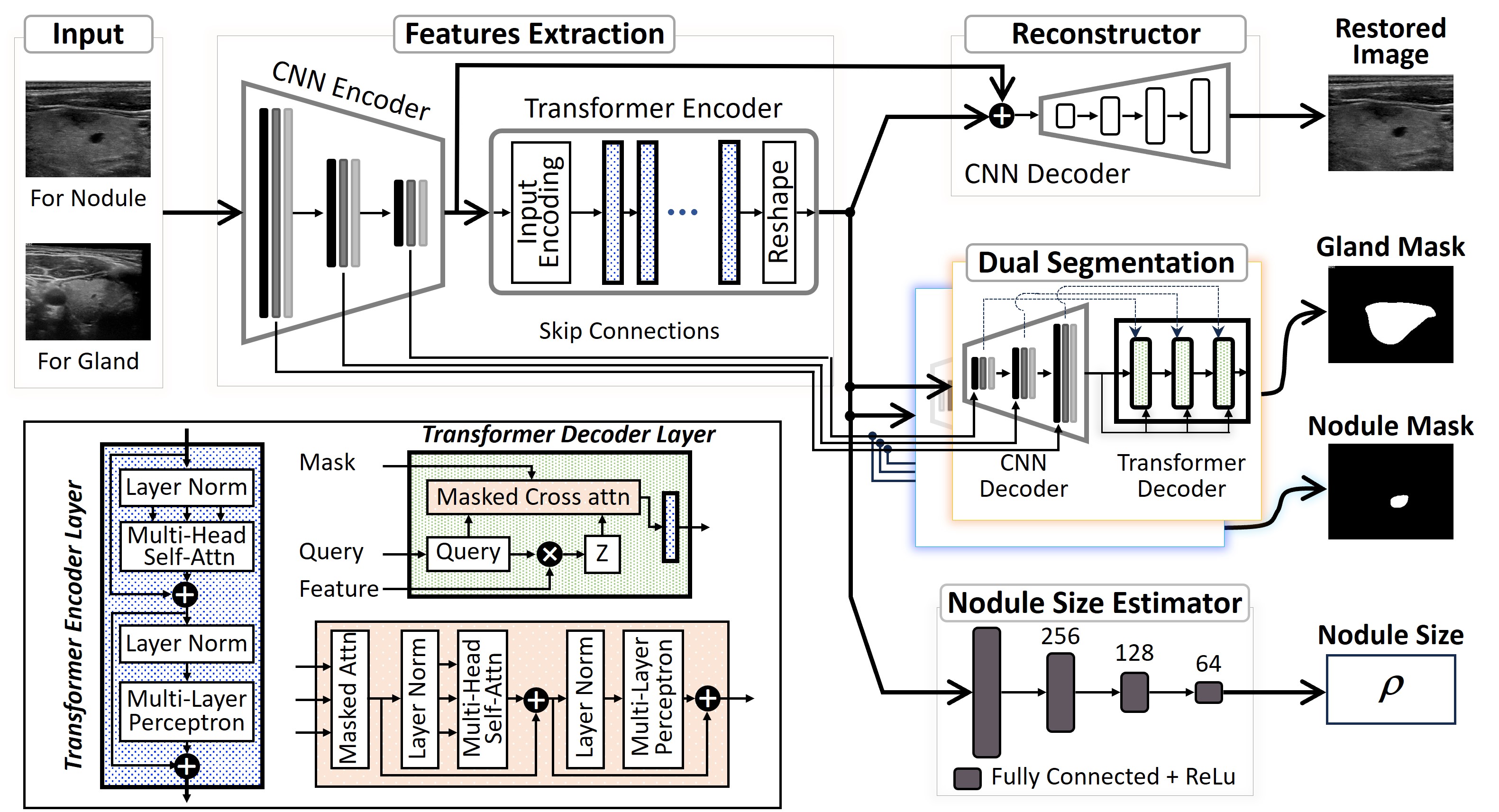}}
\caption{An overview of our SSMT-Net for thyroid nodule segmentation. }
\label{network}
\end{figure*}

The proposed SSMT-Net architecture (Fig.~\ref{network}) consists of four key modules:

\begin{enumerate}
    \item \textit{Encoder}: Extracts hierarchical features from the input thyroid ultrasound image $\mathbf{x} \in \mathbb{R}^{H \times W \times C}$, where $H \times W$ represents the spatial resolution and $C$ denotes the number of channels.

    \item \textit{Dual Segmentation}: Inspired by U-Net \cite{ronneberger2015u}, the Dual Decoder Module (DDM) comprises two parallel decoders that generate high-resolution segmentation maps for the thyroid gland and nodule. Skip connections from the encoder's CNN components are used to retain spatial details and enhance segmentation accuracy.

    \item \textit{Reconstructor}: Enforces feature consistency by reconstructing the input image, enabling unsupervised learning and improving representation learning.

    \item \textit{Nodule Size Estimator}: Predicts nodule size using encoder-extracted features as an auxiliary task, complementing segmentation by incorporating structural information.
\end{enumerate}

\subsection{Image Encoding}

Inspired by \cite{chen2021transunet,chen2024transunet}, the proposed SSMT-Net integrates Transformer layers within the encoder architecture to enhance feature representation. The input 2D image $\mathbf{x}$ is first tokenized by partitioning it into a sequence of flattened 2D patches, denoted as $\left\{\mathbf{x}^p_i \in \mathbb{R}^{P^2} \,\middle|\, i = 1, \dots, N \right\}$, where each patch has dimensions $P \times P$ and $N = \frac{HW}{P^2}$ represents the total number of patches (i.e., the input sequence length). These vectorized patches are then projected into a $d_{\mathrm{enc}}$-dimensional latent space via a trainable linear projection. To retain spatial positional information, learned position embeddings are added to the patch embeddings:

\begin{equation}
    \mathbf{z}_0 = \left[\mathbf{x}^p_1 \mathbf{E}; \mathbf{x}^p_2 \mathbf{E}; \dots; \mathbf{x}^p_N \mathbf{E}\right] + \mathbf{E}^{\mathrm{pos}},
\end{equation}
where $\mathbf{E} \in \mathbb{R}^{(P^2) \times d_{\mathrm{enc}}}$ is the patch embedding projection matrix and $\mathbf{E}^{\mathrm{pos}} \in \mathbb{R}^{d_{\mathrm{enc}}}$ represents the position embedding.

Each Transformer layer consists of a Multi-Head Self-Attention (MSA) module followed by a Multi-Layer Perceptron (MLP) block. The output at the $\ell$-th layer is computed as:

\begin{align}
    \mathbf{z}'_{\ell} &= \mathrm{MSA}\left(\mathrm{LN}\left(\mathbf{z}_{\ell-1}\right)\right) + \mathbf{z}_{\ell-1}, \\
    \mathbf{z}_{\ell}  &= \mathrm{MLP}\left(\mathrm{LN}\left(\mathbf{z}'_{\ell}\right)\right) + \mathbf{z}'_{\ell},
\end{align}
where $\mathrm{LN}(\cdot)$ denotes layer normalization and $\mathbf{z}_{\ell}$ represents the encoded image feature representation.

\subsection{Dual-Decoder Module for Segmentation} 
The features extracted from the encoder are fed into the DDM, where one decoder is responsible for gland segmentation (auxiliary task) and the other for thyroid nodule segmentation (primary task). Each decoder in the proposed dual-decoder architecture integrates both a CNN decoder and a Transformer decoder to leverage local and global contextual features, respectively. Traditional U-Net segmentation methods process each pixel independently, often leading to errors in detecting small nodules with indistinct boundaries. Instead, inspired by \cite{chen2024transunet}, our decoders formulate segmentation as a mask classification problem, where each nodule and gland is represented by a learnable query vector $\mathbf{P}_t \in \mathbb{R}^{N \times d_{\mathrm{dec}}}$, which undergoes iterative refinement through self-attention and cross-attention with multi-scale CNN features.

Given an input ultrasound image, the CNN encoder extracts a feature map $\mathbf{F} \in \mathbb{R}^{D \times H \times W \times d_{\mathrm{dec}}}$. This feature map is processed by the Transformer decoder to generate an initial coarse segmentation map:

\begin{equation}
    \mathbf{Z}_0 = g(\mathbf{P}_0 \cdot \mathbf{F}^T),
\end{equation}
where \( g(\cdot) \) represents a sigmoid activation function followed by thresholding.

At each Transformer decoder layer, nodule queries are refined using cross-attention, which updates the queries as:
\begin{equation}
    \mathbf{P}_{t+1} = \mathbf{P}_t + \text{Softmax}((\mathbf{P}_t \mathbf{W}_q)(\mathcal{F} \mathbf{W}_k)^T) \cdot (\mathcal{F} \mathbf{W}_v),
\end{equation}
where \(\mathcal{F}\) represents the projected CNN feature map in a compatible latent space, and $\mathbf{W}_q, \mathbf{W}_k$, and $\mathbf{W}_v$ are the learnable weight matrices for query, key, and value transformations respectively. Note that, for the auxiliary gland segmentation decoder, a similar refinement process is applied to gland queries.




To enhance segmentation accuracy, particularly for small nodules, we introduce a coarse-to-fine attention refinement mechanism. This mechanism restricts attention within the predicted foreground region at each iteration using a mask function, defined as:
\begin{equation}
    h(\mathbf{Z}_t(i, j, s)) =
    \begin{cases}
        0, & \text{if } \mathbf{Z}_t(i, j, s) = 1, \\
        -\infty, & \text{otherwise}.
    \end{cases}
\end{equation}
This ensures that attention is focused on the foreground region, reducing false positives and improving nodule delineation. 

After \( T \) iterations, the final refined segmentation map is computed as:
\begin{equation}
    \mathbf{Z}_T = g(\mathbf{P}_T \cdot \mathbf{F}^T).
\end{equation}

To associate each binarized mask with a semantic class, a linear classifier is applied:
\begin{equation}
    \mathbf{O} = \mathbf{P}_T \mathbf{W}_{fc},
\end{equation}
\begin{equation}
    \hat{y} = \arg \max_{k=0,1,\dots,K-1} \mathbf{O}.
\end{equation}

This dual-decoder framework effectively captures local texture details and global spatial relationships, making it well-suited for the challenging task of thyroid nodule segmentation in ultrasound imaging.

\subsection{Nodule Size Prediction} To improve the model's awareness of thyroid nodule sizes in ultrasound images, we introduce \textit{Nodule Size Prediction} as a secondary supervised auxiliary task. This task encourages the encoder to incorporate nodule size information during training, leading to more accurate segmentation. Specifically, a size prediction module is integrated after the Transformer's encoder layers. Using an MLP, the predicted nodule size \( V^{\text{pred}} \) is computed as:
\begin{equation}
    V^{\text{pred}} = \text{sigmoid}(\text{MLP}(f)),
\end{equation}
where \( f \) denotes the feature vector extracted from the Transformer's encoder, and the \textit{sigmoid} function ensures that the predicted size remains within the normalized range \([0,1]\), representing the relative area of the nodule in the image.


\subsection{Reconstruction Module} 

To incorporate unsupervised learning in SSMT-Net, we introduce a reconstruction path as an auxiliary unsupervised task. This encourages the encoder to learn robust feature representations by reconstructing the original thyroid ultrasound image. Specifically, we aggregate the final feature maps from both the CNN and Transformer encoder branches to generate the reconstructed image \( I_{\text{rec}} \):

\begin{equation}
    I_{\text{rec}} = D_{\text{REC}}(F_C + F_T),
\end{equation}
where \( D_{\text{REC}}(\cdot) \) is the reconstruction decoder, while \( F_C \) and \( F_T \) represent the final feature maps from the CNN and Transformer encoder layers, respectively.




\subsection{Semi-Supervised Multitask Framework}

The proposed SSMT-Net is designed for semi-supervised multitask learning by jointly optimizing both supervised and unsupervised tasks. The model incorporates four losses: the primary task loss for nodule segmentation (\(\mathcal{L}_{\text{nodule}}\)), two supervised auxiliary task losses for gland segmentation (\(\mathcal{L}_{\text{gland}}\)) and nodule size prediction (\(\mathcal{L}_{\text{size}}\)), and an unsupervised auxiliary loss for reconstruction (\(\mathcal{L}_{\text{rec}}\)). \( \mathcal{L}_{\text{nodule}} \) and \( \mathcal{L}_{\text{gland}} \) are based on the Dice loss. 

Training is conducted in two phases. In the first phase, the unsupervised phase, only the reconstruction task is trained using a combination of thyroid gland and nodule data while other branches remain disabled. This phase allows the model to incorporate additional unlabeled datasets, enhancing feature learning. In the second phase, the supervised phase, all tasks including segmentation, reconstruction, and nodule size prediction are jointly optimized.

To achieve balanced multitask learning, we define the total loss function as:
\begin{equation}
    \mathcal{L}_{\text{total}} = \alpha \mathcal{L}_{\text{nodule}} + \beta \mathcal{L}_{\text{gland}} + \gamma \mathcal{L}_{\text{size}} + \eta \mathcal{L}_{\text{rec}},
\end{equation}
where the parameters \( \alpha \), \( \beta \), \( \gamma \), and \( \eta \) control the contribution of each loss term, while satisfying the constraint \( \alpha + \beta + \gamma + \eta = 1 \). Since nodule segmentation is the primary task, it is assigned the highest weight, ensuring \( \alpha > \eta + \beta + \gamma \) to maintain focus. In addition, among the four losses, \( \mathcal{L}_{\text{rec}} \) is formulated using Charbonnier Loss \cite{charbonnier1994two}, an improved alternative to Mean Squared Error (MSE):
\begin{equation}
    \mathcal{L}_{\text{rec}} = \frac{1}{N} \sum_{i=1}^{N} \sqrt{(I_{\text{rec}}^i - I_{\text{input}}^i)^2 + \epsilon^2},
\end{equation}
where \( I_{\text{rec}}^i \) and \( I_{\text{input}}^i \) are the reconstructed and input images, respectively, and \( \epsilon \) (typically set to \( 10^{-6} \)) ensures numerical stability.

\section{Experiments}
\label{sec:exp&res}
\subsection{Experimental Settings} We evaluate our model on three publicly available thyroid nodule segmentation datasets: DDTI \cite{ddti_dataset}, TN3K \cite{tn3k_dataset}, and TG3K \cite{tn3k_dataset}. DDTI consists of 637 images, TN3K has 3,493 (2,879 for training, 614 for testing), and TG3K includes 3,085. Images are normalized and resized to $224 \times 224$ pixels, with augmentations such as flipping, rotation, zoom-out, and stitching. The model processes these inputs and generates segmentation masks at the same resolution.

Our model, implemented in PyTorch 2.6.0 \cite{paszke2019pytorch} and trained on an NVIDIA A40 GPU with CUDA 12.6, features a dual-encoder segmentation network combining ResNet-50 and a Vision Transformer (Base) with a $16\times16$ patch size. ResNet-50 extracts spatial features, while the Vision Transformer captures long-range dependencies. Training runs for 300 epochs with a batch size of 32 using the Adam optimizer ($LR = 0.001$, weight decay = 0.01) and a Cosine Annealing scheduler ($\text{min } LR = 1e^{-6}$).  For reproducibility, we fix the random seed to 42. We have empirically evaluated each hyperparameter and selected the optimal values: 0.8 for nodule segmentation, 0.1 for gland segmentation, 0.05 for nodule size estimation, and 0.05 for reconstruction. The model was evaluated using IoU and DSC metrics.
\subsection{Comparison with State-of-the-Art Methods}

We compare our proposed SSMT-Net with state-of-the-art thyroid nodule segmentation models, including SHAN \cite{zhang2024shan}, TRFE+ \cite{gong2023thyroid}, TnSeg \cite{ma2024tnseg}, Deblurring-MIM \cite{kang2024deblurring}, US-Net \cite{xie2024us}, and CIL-Net \cite{ali2024cil}. As shown in Table \ref{tab:sota_comparision_results}, SSMT-Net achieves the highest IoU and DSC scores on TN3K \cite{tn3k_dataset}, outperforming Deblurring-MIM by 3.32\% in IoU and TnSeg by 1.23\% in DSC, respectively. Our model surpasses CNN-based approaches by integrating both CNN and Transformer components for enhanced feature extraction and global context modeling. Unlike fully supervised methods like TRFE+, SSMT-Net employs semi-supervised learning, enabling effective feature learning with limited annotated data. Additionally, multi-task learning improves segmentation by leveraging gland segmentation for anatomical structure awareness and nodule size prediction for shape and scale refinement, resulting in more precise boundaries. These innovations collectively drive the superior performance of our approach.

\begin{table}[t]
\caption{Comparative experimental results on the TN3K \cite{tn3k_dataset} dataset.}
\centering
\resizebox{0.55\textwidth}{!}{
\begin{tabular}{|c|c|c|}
\hline
\textbf{Model} & \textbf{IoU(\%)}$\uparrow$ & \textbf{DSC(\%)}$\uparrow$ \\
\hline
TRFE+ \cite{gong2023thyroid} & 71.38 ± 0.43 & 83.30 ± 0.26 \\
SHAN \cite{zhang2024shan} & 73.59 ± 0.16 & 84.61 ± 0.04 \\
Deblurring-MIM \cite{kang2024deblurring} & 74.96 ± 0.43  & 83.68 ± 0.51 \\
Tnseg \cite{ma2024tnseg} & 73.18 ± 0.32 & 85.71 ± 0.31 \\
US-Net \cite{xie2024us} & 72.92 ± 0.38 & 83.66 ± 0.46 \\
CIL-Net \cite{ali2024cil} & 69.04 ± 0.41 & 82.14 ± 0.37 \\
SAMUS \cite{lin2024beyond} & NA & 83.05 ± NA \\
\textbf{SSMT-Net (ours)} & \textbf{78.34 ± 0.15} & \textbf{86.94 ± 0.10} \\
\hline
\end{tabular}}
\label{tab:sota_comparision_results}
\end{table}

\subsection{Ablation Study} 


Extensive experiments on the TN3K dataset \cite{tn3k_dataset} (Table \ref{tab:ablation_boundary}) assess the impact of each component on overall performance. We evaluate five model variants: (1) a baseline model without multi-task learning, based on TransUNet \cite{chen2024transunet}; (2) a model incorporating reconstruction; (3) a model adding gland segmentation; (4) a model introducing nodule size prediction; and (5) the full SSMT-Net, integrating all auxiliary tasks alongside the primary nodule segmentation task.


\begin{table}[h]
\centering
\caption{Ablation study for SSMT-Net on the TN3K \cite{tn3k_dataset} dataset.}
\resizebox{\textwidth}{!}{
\begin{tabular}{|c|c|c|c|c|c|}
\hline
\textbf{Variant} & \textbf{Reconstruction} & \textbf{Gland Segmentation} & \textbf{Nodule Size Estimation} & \textbf{IoU(\%)} & \textbf{DSC(\%)} \\
\hline
\#1 (Baseline) & $\times$ & $\times$ & $\times$ & 74.17 ± 0.21 & 82.88 ± 0.19 \\
\#2  & $\checkmark$ & $\times$ & $\times$ & 74.70 ± 0.21 & 83.26 ± 0.18 \\
\#3  & \checkmark & \checkmark & $\times$ & 75.90 ± 0.19 & 84.38 ± 0.16 \\
\#4  & \checkmark & $\times$ & \checkmark & 76.99 ± 0.17 & 85.74 ± 0.13 \\
\hline
\textbf{\#5 (All Modules)} & \checkmark & \checkmark & \checkmark & \textbf{78.34 ± 0.15} & \textbf{86.94 ± 0.10} \\
\hline
\end{tabular}}
\label{tab:ablation_boundary}
\end{table}

\begin{figure*}[!ht]
\centering
\centerline{\includegraphics[width=\textwidth]{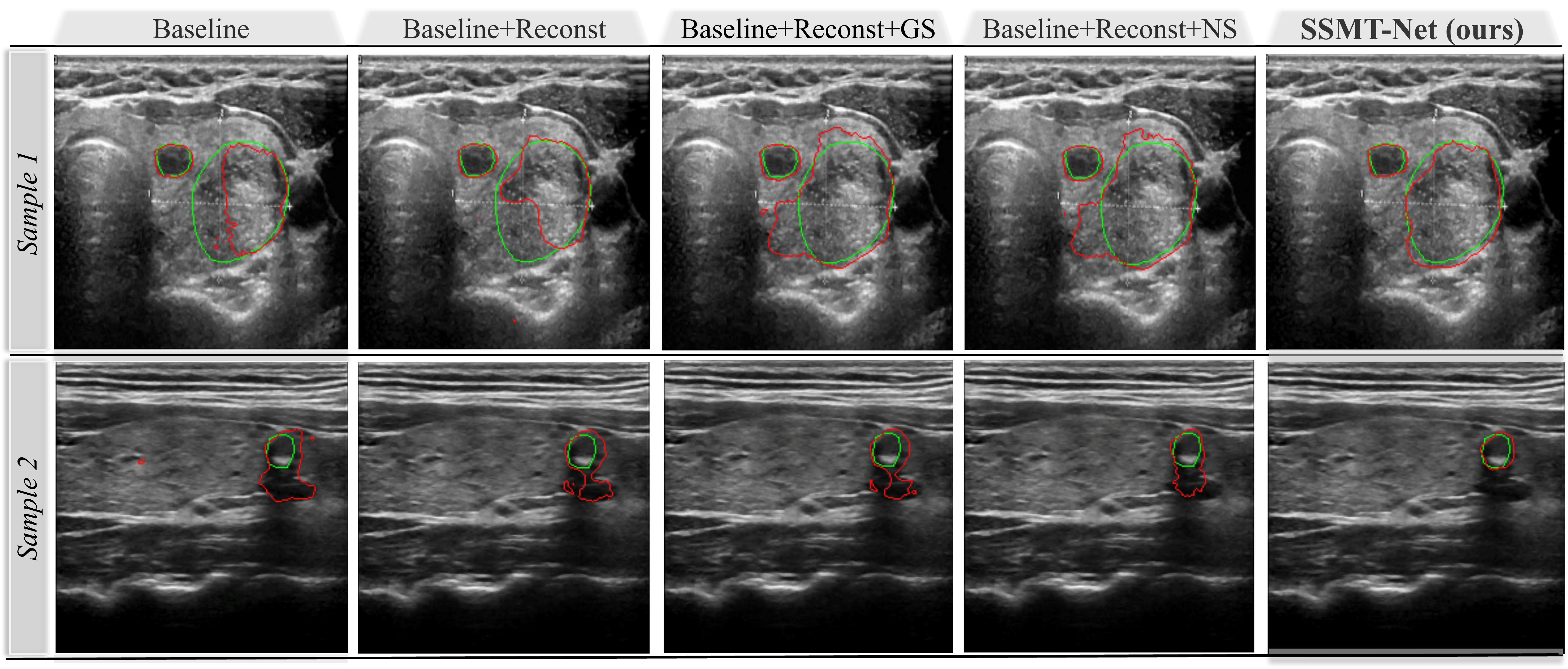}}

\caption{Visual comparison of different down-graded versions of our SSMT-Net. Original image with ground truth in green and predicted results in red.}
\label{abl_visual}
\end{figure*}

Results in Table \ref{tab:ablation_boundary} and Fig. \ref{abl_visual} show that each component improves segmentation, with optimal performance when all are combined.





\subsection{Robustness Analysis}


Table \ref{table:robust} demonstrates SSMT-Net's strong generalization in ultrasound segmentation. When tested on DDTI using a TN3K-pretrained model, performance remained competitive with state-of-the-art methods. Fine-tuning only the decoder with 20\% of DDTI while keeping the encoder frozen further improved results significantly, highlighting the model's robustness and efficient knowledge transfer. These findings confirm SSMT-Net's adaptability to domain shifts, making it ideal for real-world clinical applications with limited annotated data.


\begin{table}[t]
\centering
\caption{Robustness analysis of SSMT-Net with DDTI \cite{ddti_dataset}.}\label{table:robust}
\resizebox{0.85\textwidth}{!}{
\begin{tabular}{|c|c|c|c|c|}
\hline
\textbf{Model} & \textbf{Train set (\%)} & \textbf{Test set (\%)} & \textbf{IoU (\%)} & \textbf{DSC (\%)} \\ \hline
Tnseg \cite{ma2024tnseg}  & 90\% & 10\% & 61.11 ± 0.93 & 74.93 ± 3.02\\ \hline
DAC-Net \cite{yang2024dac}  & 80\% & 20\% & 63.20 ±  NA & 77.45 ± NA\\ \hline
Amseg \cite{ma2023amseg}  &80\% & 20\% & 60.89 ± NA & 74.81 ± NA\\ \hline
SAMUS \cite{lin2024beyond} & 20\% & 80\% & NA &  66.78 ±  NA \\ \hline
TRFE \cite{gong2021multi} & 0\% & 100\% & 52.72 ± NA & 69.04 ± NA\\ \hline
SSMT-Net (ours) & 0\% & 100\% & 56.42 ± 0.19 &  69.20 ± 0.19 \\ \hline
SSMT-Net (ours) & 20\% & 80\% & \textbf{67.68 ± 0.19} &  \textbf{78.69 ± 0.16} \\ \hline
\end{tabular}}
\end{table}


\section{Conclusions}

We proposed SSMT-Net, a semi-supervised multitask Transformer-based network for thyroid nodule segmentation, integrating four key components: an encoder, dual-decoder, reconstruction module, and nodule size predictor. The framework utilizes one unsupervised auxiliary task (reconstruction) and two supervised tasks (gland segmentation and nodule size prediction) to enhance thyroid nodule segmentation performance. Our Transformer-centric encoder and dual-decoder architecture incorporate self-attention and cross-attention mechanisms, effectively capturing both local and global dependencies. The training process consists of two phases: an unsupervised phase, where the model is pre-trained on unlabeled data via the reconstruction task, and a supervised phase, where all tasks are jointly optimized with thyroid nodule segmentation as the primary task. Ablation studies confirm that each auxiliary task improves the final segmentation performance, demonstrating the effectiveness of our multitask learning approach. The semi-supervised nature of SSMT-Net allows it to leverage unlabeled data, offering potential for further improvements and making it a strong candidate for real-world clinical applications.

\bibliographystyle{elsarticle-num}
\bibliography{refs.bib}

@article{ddti_dataset,
  title={Ultrasound image classification of thyroid nodules using machine learning techniques},
  author={Vadhiraj, Vijay Vyas and Simpkin, Andrew and O’Connell, James and Singh Ospina, Naykky and Maraka, Spyridoula and others},
  journal={Medicina},
  volume={57},
  number={6},
  pages={527},
  year={2021},
  publisher={MDPI}
}

@INPROCEEDINGS{tn3k_dataset,
  author={Gong, Haifan and Chen, Guanqi and Wang, Ranran and Xie, Xiang and Mao, Mingzhi and others},
  booktitle={2021 IEEE 18th International Symposium on Biomedical Imaging (ISBI)}, 
  title={Multi-task learning for thyroid nodule segmentation with thyroid region prior}, 
  year={2021},
  volume={},
  number={},
  pages={257-261}
}

@article{paszke2019pytorch,
  title={Pytorch: An imperative style, high-performance deep learning library},
  author={Paszke, Adam and Gross, Sam and Massa, Francisco and Lerer, Adam and Bradbury, James and Chanan, Gregory and Killeen, Trevor and Lin, Zeming and Gimelshein, Natalia and Antiga, Luca and others},
  journal={Advances in neural information processing systems},
  volume={32},
  year={2019}
}

@inproceedings{ronneberger2015u,
  title={U-net: Convolutional networks for biomedical image segmentation},
  author={Ronneberger, Olaf and Fischer, Philipp and Brox, Thomas},
  booktitle={Medical image computing and computer-assisted intervention--MICCAI 2015: 18th international conference, Munich, Germany, October 5-9, 2015, proceedings, part III 18},
  pages={234--241},
  year={2015},
  organization={Springer}
}

@article{radhachandran2024multitask,
  title={A multitask approach for automated detection and segmentation of thyroid nodules in ultrasound images},
  author={Radhachandran, Ashwath and Kinzel, Adam and Chen, Joseph and Sant, Vivek and Patel, Maitraya and Masamed, Rinat and Arnold, Corey W and Speier, William},
  journal={Computers in Biology and Medicine},
  volume={170},
  pages={107974},
  year={2024},
  publisher={Elsevier}
}

@article{gong2023thyroid,
  title={Thyroid region prior guided attention for ultrasound segmentation of thyroid nodules},
  author={Gong, Haifan and Chen, Jiaxin and Chen, Guanqi and Li, Haofeng and Li, Guanbin and Chen, Fei},
  journal={Computers in biology and medicine},
  volume={155},
  pages={106389},
  year={2023},
  publisher={Elsevier}
}

@inproceedings{gong2021multi,
  title={Multi-task learning for thyroid nodule segmentation with thyroid region prior},
  author={Gong, Haifan and Chen, Guanqi and Wang, Ranran and Xie, Xiang and Mao, Mingzhi and Yu, Yizhou and Chen, Fei and Li, Guanbin},
  booktitle={2021 IEEE 18th international symposium on biomedical imaging (ISBI)},
  pages={257--261},
  year={2021},
  organization={IEEE}
}

@inproceedings{zhang2024shan,
  title={SHAN: Shape Guided Network for Thyroid Nodule Ultrasound Cross-Domain Segmentation},
  author={Zhang, Ruixuan and Lu, Wenhuan and Guan, Cuntai and Gao, Jie and Wei, Xi and Li, Xuewei},
  booktitle={International Conference on Medical Image Computing and Computer-Assisted Intervention},
  pages={732--741},
  year={2024},
  organization={Springer}
}

@article{ma2024tnseg,
  title={Tnseg: adversarial networks with multi-scale joint loss for thyroid nodule segmentation},
  author={Ma, Xiaoxuan and Sun, Boyang and Liu, Weifeng and Sui, Dong and Shan, Sihan and Chen, Jing and Tian, Zhaofeng},
  journal={The Journal of Supercomputing},
  volume={80},
  number={5},
  pages={6093--6118},
  year={2024},
  publisher={Springer}
}

@article{kang2024deblurring,
  title={Deblurring masked image modeling for ultrasound image analysis},
  author={Kang, Qingbo and Lao, Qicheng and Gao, Jun and Liu, Jingyan and Yi, Huahui and Ma, Buyun and Zhang, Xiaofan and Li, Kang},
  journal={Medical Image Analysis},
  volume={97},
  pages={103256},
  year={2024},
  publisher={Elsevier}
}

@article{li2023novel,
  title={A novel model of thyroid nodule segmentation for ultrasound images},
  author={Li, Chengfan and Du, Ruiqi and Luo, Quanyong and Wang, Ren and Ding, Xuehai},
  journal={Ultrasound in Medicine \& Biology},
  volume={49},
  number={2},
  pages={489--496},
  year={2023},
  publisher={Elsevier}
}

@inproceedings{liu2020end,
  title={An end to end thyroid nodule segmentation model based on optimized U-net convolutional neural network},
  author={Liu, Mengya and Yuan, Xueguang and Zhang, Yangan and Chang, Kunliang and Deng, Zhifang and Xue, Jun},
  booktitle={Proceedings of the 1st International Symposium on Artificial Intelligence in Medical Sciences},
  pages={74--78},
  year={2020}
}

@article{song2024survey,
  title={A survey on deep learning in medical ultrasound imaging},
  author={Song, Ke and Feng, Jing and Chen, Duo},
  journal={Frontiers in Physics},
  volume={12},
  pages={1398393},
  year={2024},
  publisher={Frontiers Media SA}
}

@article{karkalousos2024atommic,
  title={ATOMMIC: An Advanced Toolbox for Multitask Medical Imaging Consistency to facilitate Artificial Intelligence applications from acquisition to analysis in Magnetic Resonance Imaging},
  author={Karkalousos, Dimitrios and I{\v{s}}gum, Ivana and Marquering, Henk A and Caan, Matthan WA},
  journal={Computer Methods and Programs in Biomedicine},
  pages={108377},
  year={2024},
  publisher={Elsevier}
}

@article{chen2024transunet,
  title={TransUNet: Rethinking the U-Net architecture design for medical image segmentation through the lens of transformers},
  author={Chen, Jieneng and Mei, Jieru and Li, Xianhang and Lu, Yongyi and Yu, Qihang and Wei, Qingyue and Luo, Xiangde and Xie, Yutong and Adeli, Ehsan and Wang, Yan and others},
  journal={Medical Image Analysis},
  volume={97},
  pages={103280},
  year={2024},
  publisher={Elsevier}
}

@article{xie2024us,
  title={US-Net: U-shaped network with Convolutional Attention Mechanism for ultrasound medical images},
  author={Xie, Xiaoyu and Liu, Pingping and Lang, Yijun and Guo, Zhenjie and Yang, Zhongxi and Zhao, Yuhao},
  journal={Computers \& Graphics},
  volume={124},
  pages={104054},
  year={2024},
  publisher={Elsevier}
}

@article{ali2024cil,
  title={Cil-net: Densely connected context information learning network for boosting thyroid nodule segmentation using ultrasound images},
  author={Ali, Haider and Wang, Mingzhao and Xie, Juanying},
  journal={Cognitive Computation},
  volume={16},
  number={3},
  pages={1176--1197},
  year={2024},
  publisher={Springer}
}

@inproceedings{lin2024beyond,
  title={Beyond adapting SAM: Towards end-to-end ultrasound image segmentation via auto prompting},
  author={Lin, Xian and Xiang, Yangyang and Yu, Li and Yan, Zengqiang},
  booktitle={International Conference on Medical Image Computing and Computer-Assisted Intervention},
  pages={24--34},
  year={2024},
  organization={Springer}
}

@article{chen2021transunet,
  title={Transunet: Transformers make strong encoders for medical image segmentation},
  author={Chen, Jieneng and Lu, Yongyi and Yu, Qihang and Luo, Xiangde and Adeli, Ehsan and Wang, Yan and Lu, Le and Yuille, Alan L and Zhou, Yuyin},
  journal={arXiv preprint arXiv:2102.04306},
  year={2021}
}

@article{yang2024dac,
  title={DAC-Net: A light-weight U-shaped network based efficient convolution and attention for thyroid nodule segmentation},
  author={Yang, Yingwei and Huang, Haiguang and Shao, Yingsheng and Chen, Beilei},
  journal={Computers in Biology and Medicine},
  volume={180},
  pages={108972},
  year={2024},
  publisher={Elsevier}
}

@article{ma2023amseg,
  title={Amseg: A novel adversarial architecture based multi-scale fusion framework for thyroid nodule segmentation},
  author={Ma, Xiaoxuan and Sun, Boyang and Liu, Weifeng and Sui, Dong and Chen, Jing and Tian, Zhaofeng},
  journal={IEEE Access},
  volume={11},
  pages={72911--72924},
  year={2023},
  publisher={IEEE}
}

@article{vandenhende2021multi,
  title={Multi-task learning for dense prediction tasks: A survey},
  author={Vandenhende, Simon and Georgoulis, Stamatios and Van Gansbeke, Wouter and Proesmans, Marc and Dai, Dengxin and Van Gool, Luc},
  journal={IEEE transactions on pattern analysis and machine intelligence},
  volume={44},
  number={7},
  pages={3614--3633},
  year={2021},
  publisher={IEEE}
}

@inproceedings{charbonnier1994two,
  title={Two deterministic half-quadratic regularization algorithms for computed imaging},
  author={Charbonnier, Pierre and Blanc-Feraud, Laure and Aubert, Gilles and Barlaud, Michel},
  booktitle={Proceedings of 1st international conference on image processing},
  volume={2},
  pages={168--172},
  year={1994},
  organization={IEEE}
}

@article{farooq2025gdssa,
  title={GDSSA-Net: A gradually deeply supervised self-ensemble attention network for IoMT-integrated thyroid nodule segmentation},
  author={Farooq, Muhammad Umar and Ghafoor, Haris and Rehman, Azka and Usman, Muhammad and Chae, Dong-Kyu},
  journal={Internet of Things},
  volume={31},
  pages={101598},
  year={2025},
  publisher={Elsevier}
}

@article{usman2022dual,
  title={Dual-stage deeply supervised attention-based convolutional neural networks for mandibular canal segmentation in CBCT scans},
  author={Usman, Muhammad and Rehman, Azka and Saleem, Amal Muhammad and Jawaid, Rabeea and Byon, Shi-Sub and Kim, Sung-Hyun and Lee, Byoung-Dai and Heo, Min-Suk and Shin, Yeong-Gil},
  journal={Sensors},
  volume={22},
  number={24},
  pages={9877},
  year={2022},
  publisher={MDPI}
}

@article{usman2023deha,
  title={Deha-net: A dual-encoder-based hard attention network with an adaptive roi mechanism for lung nodule segmentation},
  author={Usman, Muhammad and Shin, Yeong-Gil},
  journal={Sensors},
  volume={23},
  number={4},
  pages={1989},
  year={2023},
  publisher={MDPI}
}

@article{usman2024meds,
  title={Meds-net: Multi-encoder based self-distilled network with bidirectional maximum intensity projections fusion for lung nodule detection},
  author={Usman, Muhammad and Rehman, Azka and Shahid, Abdullah and Latif, Siddique and Shin, Yeong-Gil},
  journal={Engineering Applications of Artificial Intelligence},
  volume={129},
  pages={107597},
  year={2024},
  publisher={Elsevier}
}

@article{usman2024advancing,
  title={Advancing metaverse-based healthcare with multimodal neuroimaging fusion via multi-task adversarial variational autoencoder for brain age estimation},
  author={Usman, Muhammad and Rehman, Azka and Shahid, Abdullah and Rehman, Abd Ur and Gho, Sung-Min and Lee, Aleum and Khan, Tariq M and Razzak, Imran},
  journal={IEEE Journal of Biomedical and Health Informatics},
  year={2024},
  publisher={IEEE}
}

@article{usman2025multi,
  title={Multi-encoder self-adaptive hard attention network with maximum intensity projections for lung nodule segmentation},
  author={Usman, Muhammad and Rehman, Azka and Rehman, Abd Ur and Shahid, Abdullah and Khan, Tariq Mahmood and Razzak, Imran and Chung, Minyoung and Shin, Yeong-Gil},
  journal={Computers in Biology and Medicine},
  volume={197},
  pages={111059},
  year={2025},
  publisher={Elsevier}
}

@article{rehman2023selective,
  title={Selective deeply supervised multi-scale attention network for brain tumor segmentation},
  author={Rehman, Azka and Usman, Muhammad and Shahid, Abdullah and Latif, Siddique and Qadir, Junaid},
  journal={Sensors},
  volume={23},
  number={4},
  pages={2346},
  year={2023},
  publisher={MDPI}
}

@article{latif2020leveraging,
  title={Leveraging data science to combat COVID-19: A comprehensive review},
  author={Latif, Siddique and Usman, Muhammad and Manzoor, Sanaullah and Iqbal, Waleed and Qadir, Junaid and Tyson, Gareth and Castro, Ignacio and Razi, Adeel and Boulos, Maged N Kamel and Weller, Adrian and others},
  journal={IEEE Transactions on Artificial Intelligence},
  volume={1},
  number={1},
  pages={85--103},
  year={2020},
  publisher={IEEE}
}

@article{latif2018phonocardiographic,
  title={Phonocardiographic sensing using deep learning for abnormal heartbeat detection},
  author={Latif, Siddique and Usman, Muhammad and Rana, Rajib and Qadir, Junaid},
  journal={IEEE Sensors Journal},
  volume={18},
  number={22},
  pages={9393--9400},
  year={2018},
  publisher={IEEE}
}

@inproceedings{latif2018cross,
  title={Cross lingual speech emotion recognition: Urdu vs. western languages},
  author={Latif, Siddique and Qayyum, Adnan and Usman, Muhammad and Qadir, Junaid},
  booktitle={2018 International conference on frontiers of information technology (FIT)},
  pages={88--93},
  year={2018},
  organization={IEEE}
}

@article{usman2020volumetric,
  title={Volumetric lung nodule segmentation using adaptive ROI with multi-view residual learning},
  author={Usman, Muhammad and Lee, Byoung-Dai and Byon, Shi-Sub and Kim, Sung-Hyun and Lee, Byung-il and Shin, Yeong-Gil},
  journal={Scientific Reports},
  volume={10},
  number={1},
  pages={12839},
  year={2020},
  publisher={Nature Publishing Group UK London}
}

@article{ullah2022cascade,
  title={Cascade multiscale residual attention cnns with adaptive roi for automatic brain tumor segmentation},
  author={Ullah, Zahid and Usman, Muhammad and Jeon, Moongu and Gwak, Jeonghwan},
  journal={Information sciences},
  volume={608},
  pages={1541--1556},
  year={2022},
  publisher={Elsevier}
}

@article{usman2020retrospective,
  title={Retrospective motion correction in multishot MRI using generative adversarial network},
  author={Usman, Muhammad and Latif, Siddique and Asim, Muhammad and Lee, Byoung-Dai and Qadir, Junaid},
  journal={Scientific reports},
  volume={10},
  number={1},
  pages={4786},
  year={2020},
  publisher={Nature Publishing Group UK London}
}

@article{ullah2023densely,
  title={Densely attention mechanism based network for COVID-19 detection in chest X-rays},
  author={Ullah, Zahid and Usman, Muhammad and Latif, Siddique and Gwak, Jeonghwan},
  journal={Scientific Reports},
  volume={13},
  number={1},
  pages={261},
  year={2023},
  publisher={Nature Publishing Group UK London}
}

@article{ullah2023mtss,
  title={MTSS-AAE: Multi-task semi-supervised adversarial autoencoding for COVID-19 detection based on chest X-ray images},
  author={Ullah, Zahid and Usman, Muhammad and Gwak, Jeonghwan},
  journal={Expert Systems with Applications},
  volume={216},
  pages={119475},
  year={2023},
  publisher={Elsevier}
}

@inproceedings{usman2017using,
  title={Using deep autoencoders for facial expression recognition},
  author={Usman, Muhammad and Latif, Siddique and Qadir, Junaid},
  booktitle={2017 13th International Conference on Emerging Technologies (ICET)},
  pages={1--6},
  year={2017},
  organization={IEEE}
}

@article{ullah2023ssmd,
  title={SSMD-UNet: Semi-supervised multi-task decoders network for diabetic retinopathy segmentation},
  author={Ullah, Zahid and Usman, Muhammad and Latif, Siddique and Khan, Asifullah and Gwak, Jeonghwan},
  journal={Scientific Reports},
  volume={13},
  number={1},
  pages={9087},
  year={2023},
  publisher={Nature Publishing Group UK London}
}

@article{iqbal2023ldmres,
  title={Ldmres-net: A lightweight neural network for efficient medical image segmentation on iot and edge devices},
  author={Iqbal, Shahzaib and Khan, Tariq M and Naqvi, Syed S and Naveed, Asim and Usman, Muhammad and Khan, Haroon Ahmed and Razzak, Imran},
  journal={IEEE journal of biomedical and health informatics},
  volume={28},
  number={7},
  pages={3860--3871},
  year={2023},
  publisher={IEEE}
}

@article{lee2021evaluation,
  title={Evaluation of the feasibility of explainable computer-aided detection of cardiomegaly on chest radiographs using deep learning},
  author={Lee, Mu Sook and Kim, Yong Soo and Kim, Minki and Usman, Muhammad and Byon, Shi Sub and Kim, Sung Hyun and Lee, Byoung Il and Lee, Byoung-Dai},
  journal={Scientific reports},
  volume={11},
  number={1},
  pages={16885},
  year={2021},
  publisher={Nature Publishing Group UK London}
}

@article{latif2018automating,
  title={Automating motion correction in multishot MRI using generative adversarial networks},
  author={Latif, Siddique and Asim, Muhammad and Usman, Muhammad and Qadir, Junaid and Rana, Rajib},
  journal={arXiv preprint arXiv:1811.09750},
  year={2018}
}

@incollection{latif2018mobile,
  title={Mobile technologies for managing non-communicable diseases in developing countries},
  author={Latif, Siddique and Khan, Muhammad Yasir and Qayyum, Adnan and Qadir, Junaid and Usman, Muhammad and Ali, Syed Mustafa and Abbasi, Qammer Hussain and Imran, Muhammad Ali},
  booktitle={Mobile applications and solutions for social inclusion},
  pages={261--287},
  year={2018},
  publisher={IGI Global Scientific Publishing}
}

@article{rehman2024biological,
  title={Biological Brain Age Estimation using Sex-Aware Adversarial Variational Autoencoder with Multimodal Neuroimages},
  author={Rehman, Abd Ur and Rehman, Azka and Usman, Muhammad and Shahid, Abdullah and Gho, Sung-Min and Lee, Aleum and Khan, Tariq M and Razzak, Imran},
  journal={arXiv preprint arXiv:2412.05632},
  year={2024}
}

@article{usmancomplex,
  title={Complex Swin Transformer for Accelerating Enhanced SMWI Reconstruction},
  author={Usman, Muhammad and Gho, Sung-Min}
}

\end{document}